\documentclass[letterpaper]{article} 
\usepackage{aaai24}  
\usepackage{times}  
\usepackage{helvet}  
\usepackage{courier}  
\usepackage[hyphens]{url}  
\usepackage{graphicx} 
\usepackage{natbib}  
\usepackage{caption} 
\usepackage{algorithm}
\usepackage{algorithmic}
\usepackage{amsmath} 
\usepackage{newfloat}
\usepackage{listings}
\DeclareCaptionStyle{ruled}{labelfont=normalfont,labelsep=colon,strut=off} 
\floatstyle{ruled}
\newfloat{listing}{tb}{lst}{}
\floatname{listing}{Listing}
\title{P2CADNet: An End-to-End Network to Reconstruct Featured CAD Model from Point Cloud}
\author{
    Zhihao Zong\textsuperscript{\rm 1},
    Fazhi He\textsuperscript{\rm 1},
    Rubin Fan\textsuperscript{\rm 1},
    Yuxin Liu\textsuperscript{\rm 1}}
\affiliations{
    \textsuperscript{\rm 1} School of Computer Science,
    Wuhan University,
    China\\
    \{zhihaozong,fzhe,fanrubin,yuxinliu\}@whu.edu.cn

}

\usepackage{bibentry}

\begin{document}

\maketitle




%

\begin{abstract}

Computer Aided Design (CAD), especially the feature-based parametric CAD, plays an important role in modern industry and society. However, the reconstruction of featured CAD model is more challenging than the reconstruction of other CAD models. To this end, this paper proposes an end-to-end network to reconstruct featured CAD model from point cloud (P2CADNet). Initially, the proposed P2CADNet architecture combines a point cloud feature extractor, a CAD sequence reconstructor and a parameter optimizer. Subsequently, in order to reconstruct the featured CAD model in an autoregressive way, the CAD sequence reconstructor applies two transformer decoders, one with target mask and the other without mask. Finally, for predicting parameters more precisely, we design a parameter optimizer with cross-attention mechanism to further refine the CAD feature parameters. We evaluate P2CADNet on the public dataset, and the experimental results show that P2CADNet has excellent reconstruction quality and accuracy. To our best knowledge, P2CADNet is the first end-to-end network to reconstruct featured CAD model from point cloud, and can be regarded as baseline for future works. Therefore, we open the source code at https://github.com/Blice0415/P2CADNet.

\end{abstract}

\section{Introduction}

\begin{figure*}[htbp]
\centering
\includegraphics[width=7in]{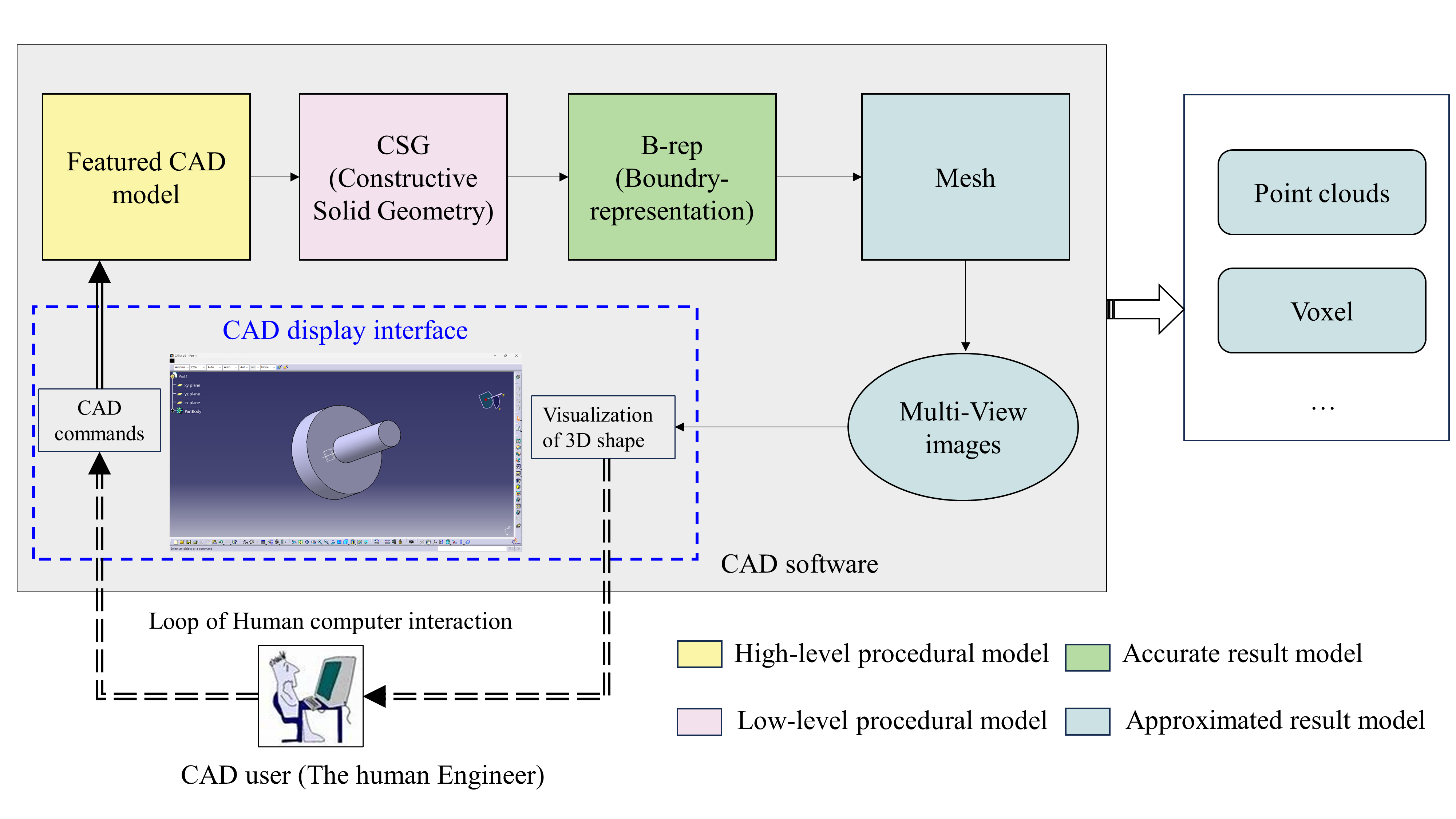}
\caption{Brief architecture of CAD and various CAD model}
\label{fig1}
\end{figure*}

CAD techniques are widely used in modern industrial design, manufacturing and other society scenarios \cite{moles2022cutting,modern,society}. Many engineers rely on CAD to construct 3D shape. However, the construction of featured CAD models costs a lot of labors and time. 

Figure \ref{fig1} shows a brief architecture of modern CAD and various CAD models. 

\textbf{Featured CAD model}. Based on loop of HCI (Human Computer Interaction), the user (CAD engineer) issues high level feature commands which include design semantics, design intentions and so on \cite{camba2016parametric,abdulla2020cad,featuredcad,featuredcadreview}. These \textbf{high level featured CAD models} are similar to natural languages and computer languages, which are easily understood and parametrically controlled by a user. To support user-centered design processing, a typical 3D shape is not directly constructed in 3D space but constructed with two steps in 2.5D space. Step 1, the user draws a sketch in 2D plane. Step 2, the user extrudes the sketch along the third 3D direction to get a full 3D shape. Furthermore, with the visualization of a 3D shape, the user can iterate the step 1 and step 2 by selecting and extruding an other 2D plane to construct a new 3D shape. In this way, the user can continue to construct more and more complex 3D shapes.

\textbf{CSG(Constructive Solid Geometry)}. In modern CAD systems, the high level featured model will be compiled and interpreted into a \textbf{low level CSG model}. Mathematically speaking, CSG is a procedural modeling approach, which combines the shape primitives with the boolean operators to construct complex shapes.



Both featured CAD models and CSG models are \textbf{procedural models }with modeling steps. Thus, it is necessary to have 
a result model to represent the final 3D shape.

\textbf{B-rep(Boundary Representation)}. As \textbf{result model} of  CSG primitives and operations, the B-rep is an accurate mathematical representation, which explicitly defines the volume limits of a 3D shape with arbitrary accuracy. 

 
\textbf{Mesh}. Mesh uses large numbers of polygons to approximate the accurate shape of B-rep. For example, one B-rep surface can be approximated with a number of triangles. 



Furthermore, mesh can also be sampled into \textbf{point cloud}, \textbf{voxel} or \textbf{multi-view images}. In a short, mesh, point cloud, voxel and multi-view images can be regarded as the discrete and approximated model for a given B-rep model. 

Both the \textbf{accurate result model} (B-rep) and the \textbf{approximated result models} (Mesh, point cloud, voxel and multi-view images) are widely used for various downstream tasks, such as visualization, CAE, CAM, CAX, VR, Animation, Simulation and so on \cite{cae,application,CAE2023}. For example, in 3D CAD systems, mesh is usually projected into multi-view images for 3D visualization in the loop of HCI by GPU rendering pipeline.


The widely used 3D datasets, such as ShapeNet \cite{shapenet}, ModelNet40 \cite{modelnet40} and ABC dataset \cite{ABC}, are derived from CAD and are claimed as CAD models. Unfortunately they generally provide result models (B-rep, mesh, point cloud, etc.) and the featured CAD model is not publicly available. However, the featured CAD model is the source of other models, and therefore is the most valuable CAD model. This is the first motivation of this paper.

Our second motivation is that a large number of 3D reconstruction researches focus on the reconstruction of result models \cite{Yin_2021_CVPR,Wen_2022_CVPR}, typically on the reconstruction of 3D mesh models \cite{meshrecon,recsurvey,10210281}. Table \ref{table1} shows the typical 3D reconstruction method with various data. Therefore, how to reconstruct featured CAD model is an open and challenging issue. 

Our third motivation is that the point cloud can be easily acquired by sensors with low cost. Therefore, automatic reconstruction of featured CAD models from point cloud will greatly save the labors and time in construction of featured CAD models.

Major contributions of this paper are as follows:

\begin{itemize}
\item For the first time, we propose an end-to-end network to directly reconstruct featured CAD sequence from point cloud without data preprocessing and intermediate data.

\item We combine both transformer decoder and masked decoder to build
a CAD sequence reconstructor, which can build CAD sequence autoregressively.


\item We adopt a cross-attention mechanism into our parameter optimizer module to predict the parameters more precisely. 

\item The proposed P2CADNet achieve the state-of-the-art performance and can be regarded as baseline for future works. 



\end{itemize}


The paper is organized as follows: Section II discusses the literature about various 3D CAD reconstructions.
The details of P2CADNet are described in Section III, and in Section IV we show the experiments and implementation details.
In Section V, we give a conclusion of this paper.

\section{Various CAD Reconstructions}

\begin{table}[tb]
    \centering
\begin{tabular}{p{2.5cm}|p{5cm}}
\hline
    Output data & Input data \\
\hline
3D mesh& image \cite{occnet}\\
& 3D point cloud \cite{occnet}\\
\hline
B-rep & mesh\cite{Breprec} \\
\hline
CSG & mesh \cite{csgnet}\\ 
\hline
Feature-based & voxel \cite{c:82}\\
Parametric CAD model& 3D point cloud \cite{c:82}\\
\hline
\end{tabular}
    \caption{3D Reconstruction Methods}
    \label{table1}
\end{table}



\subsection{Reconstruction of Mesh}

Mesh reconstruction is the most widely studied in computer vision, computer graphics, reverse engineering and so on \cite{reverse,meshrec}.

For example, mesh reconstruction from images is mostly associated with MVS(Multi-View Stereo) \cite{neuralrecon,2022CVPRmesh}. Typical methods adopt a two-stage pipeline for multi-view 3D reconstruction. The first step is to estimate the depth map for each image based on MVS. The second step is to performing depth fusion \cite{mvsnet,rmvsnet,mvsformer} to obtain the final mesh reconstruction results. 

Occupancy Networks \cite{occnet} proposes occupancy networks to reconstruct mesh from images or point cloud. It implicitly represents the 3D surface as the continuous decision boundary of a deep neural network classifier.

\subsection{Reconstruction of B-rep}

B-rep reconstruction also draws attention in computer vision, computer graphics, reverse engineering \cite{solidgen,complexgen,CVPR2023Brep}.

Re-FACE uses point cloud morphology and analysis techniques to extract features from a 3D point cloud. The authors then reconstruct the boundary contours of features using a fitting approach employing a sequence of piecewise rational Bezier curves \cite{brep2010}.

Reference \cite{brep} proposes an online approach to build planar B-Rep models from multiple organized point cloud of different viewpoints.

Product development with CAD requires analytical surfaces in B-rep. Therefore, literature \cite{Breprec} presents two methods to reconstruct B-Rep by automatic interpretation surface skeletons. One is based on decomposing inputs and the other is based on polygonal surfaces.

\subsection{Reconstruction of CSG}

CSG is a low level procedural model to construct 3D Shapes. CSG allows to combine shape primitives with boolean operators to obtain complex shapes \cite{csg2023survey}. 


Literature \cite{csg} converts the topology optimised structure into a spatial frame structure and then regenerates it in a CAD system by using standard CSG operations.  

Many other researches focus on probabilistic methods that find the most likely interpretation of the 3D shape through the inverse CSG procedure to output a CSG tree \cite{2018inversecsg,csg2021}. 

For example, CSGNet \cite{csgnet,csgnetpami} uses RNN to generate a sequence of primitives and operations  in a supervised manner and then parses the sequence as a CSG-Tree. However, annotating parsing trees for a large corpus of 3D shapes requires professional knowledge and tedious annotation processes. 

For another example, UCSG-Net\cite{ucsgnet} takes an unsupervised approach but required iterative operand selections for each tree branch.


\subsection{Reconstruction of Featured CAD Model}

Featured CAD model is a high level procedural model, which includes design semantics, design intention and so on. Therefore, the reconstruction of featured CAD model is more challenging than reconstruction of other CAD models. Thus, the related work is limited.


One related work is DeepCAD \cite{c:81}. However, it is originally proposed for generation task. In its future work, DeepCAD  uses two independent networks to reconstruct featured CAD model. The first one is to map the point cloud into a latent vector encoded by autoencoder. The second one is to decode latent vector into a CAD model by using a pretrained decoder. Therefore, DeepCAD is not an end-to-end reconstruction network. Different from DeepCAD, our network is an end-to-end network without pretrained model.

The most related work to our paper in recent years is reference \cite{c:82}, which has two steps. Firstly, it finds curves that are similar to the profile image in the 2D sketch database. Secondly, it extrudes the sketch to build a final CAD model. 

However, it \cite{c:82} has following limitations. Firstly, it heavily relies on the  third part database and traditional Dijkstra’s algorithm to search the intermediate 2D sketch. Secondly, it is not an end-to-end approach, which use both a traditional search algorithm and a extrusion decoder network. Thirdly, its source code is not publicly available.

On the contrary, the proposed P2CADNet is an end-to-end network which can directly reconstruct a featured CAD model from point cloud without intermediate 2D sketch and third part 2D sketch database. Furthermore, the source code of P2CADNet is available at https://github.com/Blice0415/P2CADNet, which can be used as baseline for future works.


\section{Proposed Method}
\begin{figure*}[htbp]
\centering
\includegraphics[width=7in]{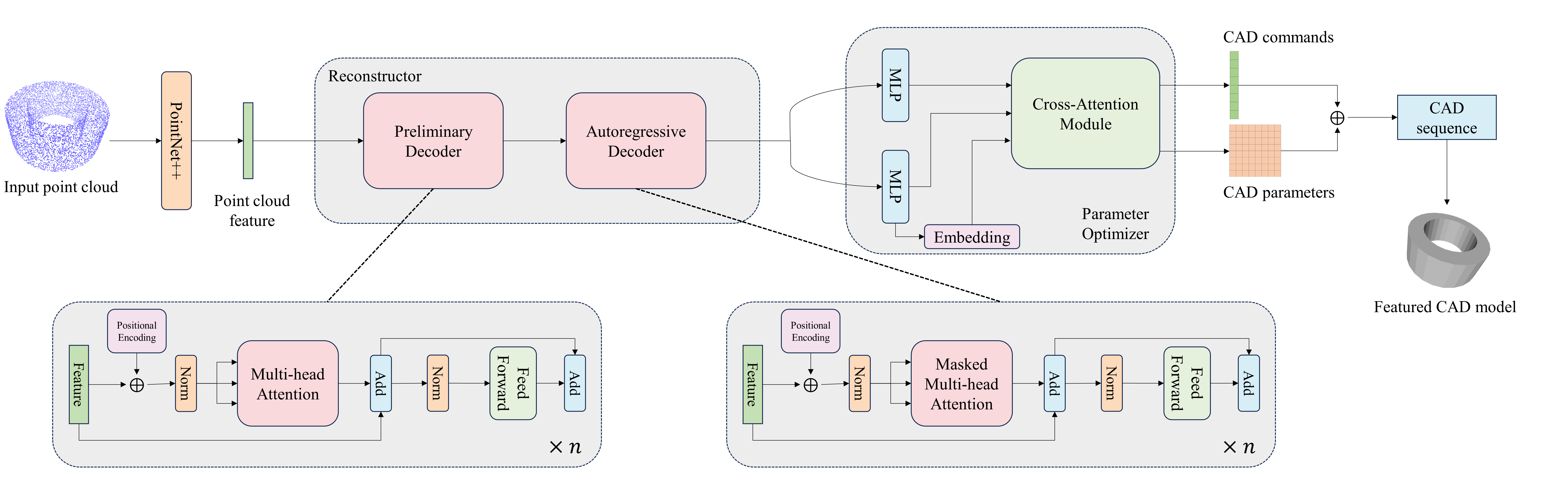}
\caption{\textbf{Our network architecture.} In general, we first input the point cloud into a point feature extractor and obtain the feature $\mathbf{F\in{{R}^{C}}}$. Then, we will pass \textbf{F} to the CAD sequence Reconstructor and output the CAD sequence feature. And the parameter Optimizer will predict the specific parameters of every featured CAD command. In the end, we can obtain the featured CAD model.}
\label{fig2}
\end{figure*}

The overview of the proposed CAD reconstruction network is shown in Figure \ref{fig2}. 

In next three subsections, we will present the proposed approach in details.

\subsection{Featured CAD sequence}

Based on mainstream literature \cite{cadfeature,cadaslanguage}, modern featured CAD model can be described as a sequence of featured CAD commands.

The CAD commands can be  2D sketch commands or 3D extrude commands. A User first creates a 2D sketch with sketch commands. And then the user applied 3D extrude commands on the sketch to build a 3D shape.

For easy discussion and description, the typical sketch commands include line, arc and circle. 
With these commands, the user can make a closed loop as sketch. Each command has the parameters to control its shape. 

The typical extrude commands include height, direction and so on. With the extrude commands, the user can extrude the 2D sketch to obtain a 3D shape. 
          

\subsection{CAD Sequence Reconstructor}

In the proposed P2CADNet, the CAD sequence reconstructor is build with transformer decoder. 

To leverage the property of CAD sequence, we apply a constant positional encoding. 
After an addition of the constant positional encoding, the input feature ${F\in{{R}^{C}}}$ will be transformed as ${F_{input}\in{{R}^{N\times C}}}$, 
\begin{equation}
F_{input} = {F + P_E}
\end{equation}
where $P_E$ is constant positional embedding. 

In the reconstructor, we combine the preliminary decoder and the autoregressive decoder.

Firstly, we use $n$ layers of transformer decoders as the preliminary decoder to output the preliminary CAD sequence features ${S\in{{R}^{N\times D}}}$. 

Secondly, we build the autoregressive decoder with $n$ layers of masked transformer decoders. The \textbf{S} will be passed to the autoregressive transformer decoder with target mask, which can autoregressively predict the CAD sequence features. 
\begin{equation}
S_{i} = \Theta({S_{0...i-1}}), i=0,1,...,N-1
\end{equation}
where $\Theta$ denotes the transformer decoders with target mask, $S_{i}$ is the i-th command of the sequence $S$.

Here, we leverage the autoregressive property of transformer decoder. After the autoregressive, we obtain the final CAD sequence features ${S\in{{R}^{N\times D}}}$.

\subsection{Parameter Optimizer} 

In order to improve the accuracy of parameter prediction, we propose a parameter optimizer based on cross-attention mechanism.

Once the reconstructed sequence features are obtained, a command as well as the parameters of the command in the sequence need to be predicted. 

An initial prediction of the command type is first obtained through a linear layer,
\begin{equation}
Cmd = {Linear(S)}, {Cmd}\in{{R}^{N\times N_{c}}}
\end{equation}

Then, we use the cross-attention mechanism to predict the command type more accurately.

\begin{equation}
Q_{cmd} = W_{Q}Cmd
\end{equation}
\begin{equation}
K_{cmd} = W_{K}S
\end{equation}
\begin{equation}
V_{cmd} = W_{V}S
\end{equation}
\begin{equation}
Cmd = Softmax(\frac{Q_{cmd}K_{cmd}^{T}V_{cmd}}{\sqrt{d^K}})
\end{equation}
Where $W_Q\in{{{R}^{N_{c}\times d}}}$,$W_K\in{{{R}^{D\times d}}}$,
$W_V\in{{{R}^{D\times N_{c}}}}$,
$Cmd\in{{{R}^{N\times N_{c}}}}$.

Finally, the parameters for every command $Param$ are predicted by a linear layer,
\begin{equation}
Param = {Linear(S)}, {Param}\in{{R}^{N\times n_{param}}}
\end{equation}

Different from the prediction of command type in equations (4-7), the prediction of parameters needs the existing command type information and global information according equations (9-12).

\begin{equation}
Q_{param} = W_{Q}Param
\end{equation}
\begin{equation}
K_{param} = W_{K}(S+Embedding(Cmd))
\end{equation}
\begin{equation}
V_{param} = W_{V}(S+Embedding(Cmd))
\end{equation}
\begin{equation}
Param = Softmax(\frac{Q_{param}K_{param}^{T}V_{param}}{\sqrt{d^K}})
\end{equation}
Where $W_Q\in{{{R}^{n_{param}\times d}}}$,$W_K\in{{{R}^{D\times d}}}$,
$W_V\in{{{R}^{D\times n_{param}}}}$,
$Param\in{{{R}^{N\times n_{param}}}}$.

To aggregate command type information and global information, we embed the command type into a embedding space and apply cross-attention on it.

\subsection{Loss Function}

To train our P2CADNet, we combine two loss functions: command loss and parameter loss.

Specifically, the command is a 1-D vector, and  therefore the command loss is defined as follows,

\begin{equation}
\mathcal{L}_{cmd} = \sum\limits_{i=1}^{N_c}l(\hat{c}_i,c_i)
\end{equation}
where $l(\cdot,\cdot)$ denotes the standard Cross-Entropy, $N_c$ is the length of sequence. $\hat{c}_i$ and $c_i$ is the ground truth commands and predicted commands. 

In addition, considering that the parameter is a 2-D vector,  we define the parameter loss as follows, 

\begin{equation}
\mathcal{L}_{param} = \sum\limits_{i=1}^{N_c}\sum\limits_{j=1}^{N_p}l(\hat{p}_{i,j},p_{i,j})
\end{equation}
where $N_p$ is the number of parameters, $\hat{p}_{i,j}$ and $p_{i,j}$ is the ground truth parameters and predicted parameters.

Finally, the overall loss function is defined as follows,

\begin{equation}
\mathcal{L} = \mathcal{L}_{cmd} + \beta\mathcal{L}_{param}
\end{equation}
where $\beta$ is the hyper-parameter.

\section{Experiment}

\subsection{Dataset}

As discussed in related work, two references \cite{c:81,c:82} are closed to our work although the P2CADNet is first end-to-end reconstruction network for featured CAD model
from point cloud. 

Thus, for fair comparison, we adopt the same dataset as that used in references \cite{c:81,c:82}, which consist of 178, 238 CAD command sequences and each CAD model can be sampled to point cloud with 2048 points.

\subsection{Metrics}

To evaluate our method, we use four metrics in our experiment. (1) Command Accuracy(ACC$_{cmd}$), (2) Parameter Accuracy(ACC$_{param}$), (3) Chamfer Distance(CD), (4) Invalid Ratio.

\subsubsection{Command Accuracy} To evaluate the accuracy of the reconstructed CAD sequence, we use Command Accuracy(ACC$_{cmd}$) to measure the command. It is defined as,
\begin{equation}
ACC_{cmd} = \frac{1}{N_c}\sum\limits_{i=1}^{N_c}\mathcal{I}[\hat{c}_i,c_i]
\end{equation}
where $N_c$ is the length of sequence, $\hat{c}_i$ is the ground truth command type, $c_i$ is the predicted command type, $\mathcal{I}$ denotes the indicator function.
\subsubsection{Parameter Accuracy} To evaluate the accuracy of the parameters, we use Parameter Accuracy(ACC$_{param}$), which is defined as,
\begin{equation}
ACC_{param} =  \frac{1}{K}\sum\limits_{i=1}^{N_c}\sum\limits_{j=1}^{N_p}\mathcal{I}[\left\vert\hat{p}_{i,j}-p_{i,j}\right\vert < \eta]\mathcal{I}[c_i = \hat{c}_i]
\end{equation}
where $\eta$ is the threshold value of the error, and we only evaluated parameters of correct command.
\subsubsection{Chamfer Distance}
Chamfer Distance(CD) is widely used in point cloud reconstruction evaluating \cite{chamfer2,chamfer1}. It is the average shortest point distance between the reconstructed point cloud and the ground truth point cloud. It is defined as, 

\begin{equation}
\begin{aligned}
CD(S_1,S_2) = \frac{1}{\left\vert S_1 \right\vert} \sum_{x\in S_1} \min_{y\in S_2} \left \|x-y\right \|_2 \\
+ \frac{1}{\left\vert S_2 \right\vert} \sum_{y\in S_2} \min_{x\in S_1} \left \|y-x\right \|_2
\end{aligned}
\end{equation}
where $S_1$ and $S_2$ denotes the reconstructed point cloud and the ground truth point cloud, $\left\vert S_1 \right\vert$ and  $\left\vert S_2 \right\vert$ denotes the number of points.
\subsubsection{Invalid Ratio}
In some cases, the output CAD sequence may lead to an invalid topology. Thus, invalid ratio is used to measure this situation.

\subsection{Implementation Details}

All experiments are conducted on a PC with the Intel I7 CPU and NVIDIA GeForce RTX 3090 GPU. 

The hidden dimension $d_{model}$ is setting to 256. The feedforward dimension is setting to 1024. $N_c$ is setting as 60 and $N_p$ is setting as 16. 

Decoder layers and masked
decoder layers are both setting to 4. We train the network for 300 epochs with learning rate 0.0001 and batch size 32.

\subsection{Comparison}

Among the two closed works, only DeepCAD \cite{c:81} provides source code. Thus, we compare P2CADNet with DeepCAD.  

For a fair comparison, we use the same experiment settings and environment as that in DeepCAD. And we run the source codes of both P2CADNet and DeepCAD in the same environment.


Table \ref{tableresult} shows the quantitative results. The bold results are the best. As Table \ref{tableresult} shows, P2CADNet achieveS more precisely commands and parameters. 

Furthermore, P2CADNet significantly reduces the median chamfer distance, which demonstrates the excellent reconstruction quality of our method. 

\begin{table}[htbp]
    \centering
    \resizebox{3.3in}{!}{
    \begin{tabular}{c c c c c}
        \hline
        Method & ACC$_{cmd} \uparrow$ & ACC$_{param}\uparrow$ & median & Invalid  \\
        & & &  CD $\downarrow$ & Ratio$\downarrow$\\
        \hline
        DeepCAD &  77.5 & 70.9 & 11.02 & \textbf{15.97}\\
        Ours & \textbf{80.6} & \textbf{75.6} & \textbf{1.90} & 23.00\\
        \hline
    \end{tabular}
    }
    \caption{Quantitative Result. ACC$_{cmd}$ and ACC$_{param}$ are both multiplied by $100\%$, and CD is multiplied by $10^3$.}
    \label{tableresult}
\end{table}

\subsection{Limitation}

The limitation of our method is that the invalid ratio is slightly higher than that of DeepCAD. 

And the reason for this, we believe that it is due to the trade-off made by our method, in which P2CADNet achieves an overwhelming reconstruction quality - the median chamfer distance.


\begin{table}[htbp]
    \centering
    \resizebox{3.3in}{!}{
    \begin{tabular}{c c c c c}
        \hline
        Method & ACC$_{cmd} \uparrow$ & ACC$_{param}\uparrow$ & median & Invalid  \\
        & & &  CD $\downarrow$ & Ratio$\downarrow$\\
        \hline
        P2CADNet w/o Mask&  80.1 & 75.0 & 2.11 & 26.61\\
        P2CADNet w/o PO & 79.7  & 74.8  & 2.02 & 24.64\\
        P2CADNet & \textbf{80.6} & \textbf{75.6} & \textbf{1.90} & \textbf{23.00}\\
        \hline
    \end{tabular}
    }
    \caption{Ablation Study. ACC$_{cmd}$ and ACC$_{param}$ are both multiplied by $100\%$, and CD is multiplied by $10^3$.}
    \label{tableablation}
\end{table}

\subsection{Ablation Study}

To evaluate the effectiveness of our proposed modules in P2CADNet, we conduct ablation studies on different settings. 

In Table \ref{tableablation}, the P2CADNet w/o Mask denotes the architecture without masked autoregressive decoder module, while the P2CADNet w/o PO denotes the architecture without parameter optimizer module.

As shown in Table \ref{tableablation}, our proposed modules improve all the metrics in experiments, which demonstrates the effectiveness of the proposed method. 

Firstly, the masked decoder module enhances the reconstruct quality (Chamfer Distance) from 2.11 to 1.90 and invalid ratio from 26.61\% to 23.00\%. 

Secondly, the parameter optimizer module enhances the accuracy of the commands from 79.7 to 80.6 and the accuracy of the parameters from 74.8 to 75.6. 

Finally, when combining two proposed modules, P2CADNet can achieve the best result.

\section{Conclusion}
Among various CAD models, the featured CAD model is the most valuable model. However, the reconstruction of a featured CAD model is open issue and is more challenging than the reconstruction of other CAD models. 

In this paper, we propose a novel end-to-end network P2CADNet to reconstruct featured CAD models from point cloud. Specifically, we propose a CAD sequence reconstructor, which combines the transformer decoder layers and the masked transformer decoder layers to autoregressively reconstruct featured CAD sequence. We further design a parameter optimizer based on cross-attention mechanism to precisely predict the parameters.
Finally, we conduct experiments on public dataset to demonstrate the effectiveness of P2CADNet.

To our best knowledge, P2CADNet is the first end-to-end network to reconstruct featured CAD model from point cloud. Furthermore, our source code is available at https://github.com/Blice0415/P2CADNet, which can be regarded as baseline for future works.


\bibliography{aaai24}

\begin{thebibliography}{45}
\providecommand{\natexlab}[1]{#1}

\bibitem[{Abdulla, Ali, and Jamel(2020)}]{abdulla2020cad}
Abdulla, M.~A.; Ali, H.; and Jamel, R.~S. 2020.
\newblock CAD-CAM technology: a literature review.
\newblock \emph{Al-Rafidain Dental Journal}, 20(1): 95--113.

\bibitem[{Camba, Contero, and Company(2016)}]{camba2016parametric}
Camba, J.~D.; Contero, M.; and Company, P. 2016.
\newblock Parametric CAD modeling: An analysis of strategies for design
  reusability.
\newblock \emph{Computer-Aided Design}, 74: 18--31.

\bibitem[{Camba, Hartman, and Bertoline(2023)}]{modern}
Camba, J.~D.; Hartman, N.; and Bertoline, G.~R. 2023.
\newblock Computer-Aided Design, Computer-Aided Engineering, and Visualization.
\newblock In \emph{Springer Handbook of Automation}, 641--659. Springer.

\bibitem[{Cao, Ren, and Fu(2022, online)}]{mvsformer}
Cao, C.; Ren, X.; and Fu, Y. 2022, online.
\newblock {MVSF}ormer: Multi-View Stereo by Learning Robust Image Features and
  Temperature-based Depth.
\newblock \emph{Transactions on Machine Learning Research}.

\bibitem[{Chang et~al.(2015)Chang, Funkhouser, Guibas, Hanrahan, Huang, Li,
  Savarese, Savva, Song, Su, Xiao, Yi, and Yu}]{shapenet}
Chang, A.~X.; Funkhouser, T.; Guibas, L.; Hanrahan, P.; Huang, Q.; Li, Z.;
  Savarese, S.; Savva, M.; Song, S.; Su, H.; Xiao, J.; Yi, L.; and Yu, F. 2015.
\newblock {ShapeNet: An Information-Rich 3D Model Repository}.
\newblock Technical Report arXiv:1512.03012 [cs.GR], Stanford University ---
  Princeton University --- Toyota Technological Institute at Chicago.

\bibitem[{Du et~al.(2018)Du, Inala, Pu, Spielberg, Schulz, Rus, Solar-Lezama,
  and Matusik}]{2018inversecsg}
Du, T.; Inala, J.~P.; Pu, Y.; Spielberg, A.; Schulz, A.; Rus, D.; Solar-Lezama,
  A.; and Matusik, W. 2018.
\newblock Inversecsg: Automatic conversion of 3d models to csg trees.
\newblock \emph{ACM Transactions on Graphics (TOG)}, 37(6): 1--16.

\bibitem[{Fan, Su, and Guibas(2017)}]{chamfer2}
Fan, H.; Su, H.; and Guibas, L.~J. 2017.
\newblock A Point Set Generation Network for 3D Object Reconstruction From a
  Single Image.
\newblock In \emph{Proceedings of the IEEE Conference on Computer Vision and
  Pattern Recognition (CVPR)}, 605--613.

\bibitem[{Fayolle and Friedrich(2023)}]{csg2023survey}
Fayolle, P.-A.; and Friedrich, M. 2023.
\newblock A Survey of Methods for Converting Unstructured Data to CSG Models.
\newblock \emph{arXiv preprint arXiv:2305.01220}.

\bibitem[{Ganin et~al.(2021)Ganin, Bartunov, Li, Keller, and
  Saliceti}]{cadaslanguage}
Ganin, Y.; Bartunov, S.; Li, Y.; Keller, E.; and Saliceti, S. 2021.
\newblock Computer-aided design as language.
\newblock \emph{Advances in Neural Information Processing Systems}, 34:
  5885--5897.

\bibitem[{Guo et~al.(2022)Guo, Liu, Pan, Liu, Tong, and Guo}]{complexgen}
Guo, H.; Liu, S.; Pan, H.; Liu, Y.; Tong, X.; and Guo, B. 2022.
\newblock Complexgen: Cad reconstruction by b-rep chain complex generation.
\newblock \emph{ACM Transactions on Graphics (TOG)}, 41(4): 1--18.

\bibitem[{Huang et~al.(2022)Huang, Wen, Wang, Ren, and Jia}]{recsurvey}
Huang, Z.; Wen, Y.; Wang, Z.; Ren, J.; and Jia, K. 2022.
\newblock Surface Reconstruction from Point Clouds: A Survey and a Benchmark.
\newblock arXiv:2205.02413.

\bibitem[{Jayaraman et~al.(2022)Jayaraman, Lambourne, Desai, Willis, Sanghi,
  and Morris}]{solidgen}
Jayaraman, P.~K.; Lambourne, J.~G.; Desai, N.; Willis, K.~D.; Sanghi, A.; and
  Morris, N.~J. 2022.
\newblock Solidgen: An autoregressive model for direct b-rep synthesis.
\newblock \emph{arXiv preprint arXiv:2203.13944}.

\bibitem[{Jones et~al.(2023)Jones, Hu, Kodnongbua, Kim, and
  Schulz}]{CVPR2023Brep}
Jones, B.~T.; Hu, M.; Kodnongbua, M.; Kim, V.~G.; and Schulz, A. 2023.
\newblock Self-Supervised Representation Learning for CAD.
\newblock In \emph{Proceedings of the IEEE/CVF Conference on Computer Vision
  and Pattern Recognition (CVPR)}, 21327--21336.

\bibitem[{Kania, Zieba, and Kajdanowicz(2020)}]{ucsgnet}
Kania, K.; Zieba, M.; and Kajdanowicz, T. 2020.
\newblock UCSG-NET-unsupervised discovering of constructive solid geometry
  tree.
\newblock \emph{Advances in Neural Information Processing Systems}, 33:
  8776--8786.

\bibitem[{Koch et~al.(2019)Koch, Matveev, Jiang, Williams, Artemov, Burnaev,
  Alexa, Zorin, and Panozzo}]{ABC}
Koch, S.; Matveev, A.; Jiang, Z.; Williams, F.; Artemov, A.; Burnaev, E.;
  Alexa, M.; Zorin, D.; and Panozzo, D. 2019.
\newblock ABC: A Big CAD Model Dataset for Geometric Deep Learning.
\newblock In \emph{Proceedings of the IEEE/CVF Conference on Computer Vision
  and Pattern Recognition (CVPR)}, 9601--9611.

\bibitem[{Kyratsis, Kakoulis, and Markopoulos(2020)}]{cae}
Kyratsis, P.; Kakoulis, K.; and Markopoulos, A.~P. 2020.
\newblock Advances in CAD/CAM/CAE Technologies.
\newblock \emph{Machines}, 8(1).

\bibitem[{Lambourne et~al.(2022)Lambourne, Willis, Jayaraman, Zhang, Sanghi,
  and Malekshan}]{c:82}
Lambourne, J.~G.; Willis, K.; Jayaraman, P.~K.; Zhang, L.; Sanghi, A.; and
  Malekshan, K.~R. 2022.
\newblock Reconstructing Editable Prismatic CAD from Rounded Voxel Models.
\newblock In \emph{SIGGRAPH Asia 2022 Conference Papers}, SA '22. New York, NY,
  USA: Association for Computing Machinery.
\newblock ISBN 9781450394703.

\bibitem[{Li et~al.(2023{\natexlab{a}})Li, Lin, Chen, Liu, Gao, and
  Zou}]{CAE2023}
Li, M.; Lin, C.; Chen, W.; Liu, Y.; Gao, S.; and Zou, Q. 2023{\natexlab{a}}.
\newblock XVoxel-Based Parametric Design Optimization of Feature Models.
\newblock \emph{Computer-Aided Design}, 160: 103528.

\bibitem[{Li et~al.(2023{\natexlab{b}})Li, Hu, Ouyang, and Shen}]{10210281}
Li, Y.; Hu, Q.; Ouyang, Z.; and Shen, S. 2023{\natexlab{b}}.
\newblock Neural Reflectance Decomposition Under Dynamic Point Light.
\newblock \emph{IEEE Transactions on Circuits and Systems for Video
  Technology}, 1--1.

\bibitem[{Lin, Kong, and Lucey(2018)}]{chamfer1}
Lin, C.-H.; Kong, C.; and Lucey, S. 2018.
\newblock Learning Efficient Point Cloud Generation for Dense 3D Object
  Reconstruction.
\newblock \emph{Proceedings of the AAAI Conference on Artificial Intelligence},
  32(1).

\bibitem[{Mayer et~al.(2022)Mayer, V{\"o}lkl, Wartzack et~al.}]{Breprec}
Mayer, J.; V{\"o}lkl, H.; Wartzack, S.; et~al. 2022.
\newblock Feature-Based Reconstruction of Non-Beam-Like Topology Optimization
  Design Proposals in Boundary-Representation.
\newblock In \emph{DS 119: Proceedings of the 33rd Symposium Design for X
  (DFX2022)}, 1--10.

\bibitem[{Mescheder et~al.(2019)Mescheder, Oechsle, Niemeyer, Nowozin, and
  Geiger}]{occnet}
Mescheder, L.; Oechsle, M.; Niemeyer, M.; Nowozin, S.; and Geiger, A. 2019.
\newblock Occupancy Networks: Learning 3D Reconstruction in Function Space.
\newblock In \emph{Proceedings of the IEEE/CVF Conference on Computer Vision
  and Pattern Recognition (CVPR)}, 4460--4470.

\bibitem[{Moles et~al.(2022)Moles, Suhaym, Palla, and
  Callahan}]{moles2022cutting}
Moles, S.~L.; Suhaym, O.; Palla, B.~L.; and Callahan, N.~F. 2022.
\newblock Cutting guides in Mandibular Tumor Ablation: Are We as Accurate as We
  Think?
\newblock \emph{Journal of Oral and Maxillofacial Surgery}, 80(9): S73.

\bibitem[{Niu et~al.(2015)Niu, Martin, Langbein, and Sabin}]{cadfeature}
Niu, Z.; Martin, R.~R.; Langbein, F.~C.; and Sabin, M.~A. 2015.
\newblock Rapidly finding CAD features using database optimization.
\newblock \emph{Computer-Aided Design}, 69: 35--50.

\bibitem[{Otto and Mandorli(2023)}]{society}
Otto, H.~E.; and Mandorli, F. 2023.
\newblock Graphical Representation of Parametric Feature-Based MCAD Model
  Characteristics.
\newblock \emph{Computer-Aided Design and Applications}, 20(2).

\bibitem[{Pang et~al.(2023)Pang, Peng, Dong, Yuan, Wang, and Sun}]{meshrec}
Pang, S.; Peng, R.; Dong, Y.; Yuan, Q.; Wang, S.; and Sun, J. 2023.
\newblock JointMETRO: a 3D reconstruction model for human figures in works of
  art based on transformer.
\newblock \emph{Neural Computing and Applications}, 1--15.

\bibitem[{Raffo, Barrowclough, and Muntingh(2020)}]{reverse}
Raffo, A.; Barrowclough, O.~J.; and Muntingh, G. 2020.
\newblock Reverse engineering of CAD models via clustering and approximate
  implicitization.
\newblock \emph{Computer Aided Geometric Design}, 80: 101876.

\bibitem[{Ren et~al.(2021)Ren, Zheng, Cai, Li, Jiang, Cai, Zhang, Pan, Zhang,
  Zhao et~al.}]{csg2021}
Ren, D.; Zheng, J.; Cai, J.; Li, J.; Jiang, H.; Cai, Z.; Zhang, J.; Pan, L.;
  Zhang, M.; Zhao, H.; et~al. 2021.
\newblock Csg-stump: A learning friendly csg-like representation for
  interpretable shape parsing.
\newblock In \emph{Proceedings of the IEEE/CVF International Conference on
  Computer Vision}, 12478--12487.

\bibitem[{Sand and Henrich(2016)}]{brep}
Sand, M.; and Henrich, D. 2016.
\newblock Incremental reconstruction of planar B-Rep models from multiple point
  clouds.
\newblock \emph{The Visual Computer: International Journal of Computer
  Graphics}, 32(6-8): 945--954.

\bibitem[{Sepp{\"a}l{\"a} et~al.(2022)Sepp{\"a}l{\"a}, Saukkoriipi, Lohi,
  Soutukorva, Heikkil{\"a}, and Koskinen}]{application}
Sepp{\"a}l{\"a}, T.; Saukkoriipi, J.; Lohi, T.; Soutukorva, S.; Heikkil{\"a},
  T.; and Koskinen, J. 2022.
\newblock Feature-Based Object Detection and Pose Estimation Based on 3D
  Cameras and CAD Models for Industrial Robot Applications.
\newblock In \emph{2022 18th IEEE/ASME International Conference on Mechatronic
  and Embedded Systems and Applications (MESA)}, 1--5. IEEE.

\bibitem[{Sharma et~al.(2018)Sharma, Goyal, Liu, Kalogerakis, and
  Maji}]{csgnet}
Sharma, G.; Goyal, R.; Liu, D.; Kalogerakis, E.; and Maji, S. 2018.
\newblock Csgnet: Neural shape parser for constructive solid geometry.
\newblock In \emph{Proceedings of the IEEE Conference on Computer Vision and
  Pattern Recognition}, 5515--5523.

\bibitem[{Sharma et~al.(2022)Sharma, Goyal, Liu, Kalogerakis, and
  Maji}]{csgnetpami}
Sharma, G.; Goyal, R.; Liu, D.; Kalogerakis, E.; and Maji, S. 2022.
\newblock Neural Shape Parsers for Constructive Solid Geometry.
\newblock \emph{IEEE Transactions on Pattern Analysis and Machine
  Intelligence}, 44(5): 2628--2640.

\bibitem[{{\v{S}}klebar et~al.(2023){\v{S}}klebar, Martinec, Peri{\v{s}}i{\'c},
  and {\v{S}}torga}]{featuredcadreview}
{\v{S}}klebar, J.; Martinec, T.; Peri{\v{s}}i{\'c}, M.~M.; and {\v{S}}torga, M.
  2023.
\newblock CLUSTERING OF SEQUENTIAL CAD MODELLING DATA.
\newblock \emph{Proceedings of the Design Society}, 3: 937--946.

\bibitem[{Stamati and Fudos(2010)}]{brep2010}
Stamati, V.; and Fudos, I. 2010.
\newblock Building editable B-Rep models from unorganized point clouds.
\newblock \emph{Jul}, 7: 1--10.

\bibitem[{Sun et~al.(2021)Sun, Xie, Chen, Zhou, and Bao}]{neuralrecon}
Sun, J.; Xie, Y.; Chen, L.; Zhou, X.; and Bao, H. 2021.
\newblock NeuralRecon: Real-time coherent 3D reconstruction from monocular
  video.
\newblock In \emph{Proceedings of the IEEE/CVF Conference on Computer Vision
  and Pattern Recognition}, 15598--15607.

\bibitem[{Wang et~al.(2021)Wang, Zhang, Li, Fu, Yu, Liu, Xue, and
  Jiang}]{meshrecon}
Wang, N.; Zhang, Y.; Li, Z.; Fu, Y.; Yu, H.; Liu, W.; Xue, X.; and Jiang, Y.-G.
  2021.
\newblock Pixel2Mesh: 3D Mesh Model Generation via Image Guided Deformation.
\newblock \emph{IEEE Transactions on Pattern Analysis and Machine
  Intelligence}, 43(10): 3600--3613.

\bibitem[{Wen et~al.(2022)Wen, Zhou, Liu, Su, Dong, and Han}]{Wen_2022_CVPR}
Wen, X.; Zhou, J.; Liu, Y.-S.; Su, H.; Dong, Z.; and Han, Z. 2022.
\newblock 3D Shape Reconstruction From 2D Images With Disentangled Attribute
  Flow.
\newblock In \emph{Proceedings of the IEEE/CVF Conference on Computer Vision
  and Pattern Recognition (CVPR)}, 3803--3813.

\bibitem[{Worchel et~al.(2022)Worchel, Diaz, Hu, Schreer, Feldmann, and
  Eisert}]{2022CVPRmesh}
Worchel, M.; Diaz, R.; Hu, W.; Schreer, O.; Feldmann, I.; and Eisert, P. 2022.
\newblock Multi-View Mesh Reconstruction With Neural Deferred Shading.
\newblock In \emph{Proceedings of the IEEE/CVF Conference on Computer Vision
  and Pattern Recognition (CVPR)}, 6187--6197.

\bibitem[{Wu, Xiao, and Zheng(2021)}]{c:81}
Wu, R.; Xiao, C.; and Zheng, C. 2021.
\newblock DeepCAD: A Deep Generative Network for Computer-Aided Design Models.
\newblock In \emph{Proceedings of the IEEE/CVF International Conference on
  Computer Vision (ICCV)}, 6772--6782.

\bibitem[{Wu et~al.(2015)Wu, Song, Khosla, Yu, Zhang, Tang, and
  Xiao}]{modelnet40}
Wu, Z.; Song, S.; Khosla, A.; Yu, F.; Zhang, L.; Tang, X.; and Xiao, J. 2015.
\newblock 3d shapenets: A deep representation for volumetric shapes.
\newblock In \emph{Proceedings of the IEEE conference on computer vision and
  pattern recognition}, 1912--1920.

\bibitem[{Yao et~al.(2018)Yao, Luo, Li, Fang, and Quan}]{mvsnet}
Yao, Y.; Luo, Z.; Li, S.; Fang, T.; and Quan, L. 2018.
\newblock Mvsnet: Depth inference for unstructured multi-view stereo.
\newblock In \emph{Proceedings of the European conference on computer vision
  (ECCV)}, 767--783.

\bibitem[{Yao et~al.(2019)Yao, Luo, Li, Shen, Fang, and Quan}]{rmvsnet}
Yao, Y.; Luo, Z.; Li, S.; Shen, T.; Fang, T.; and Quan, L. 2019.
\newblock Recurrent mvsnet for high-resolution multi-view stereo depth
  inference.
\newblock In \emph{Proceedings of the IEEE/CVF conference on computer vision
  and pattern recognition}, 5525--5534.

\bibitem[{Yin, Xiao, and Cirak(2020)}]{csg}
Yin, G.; Xiao, X.; and Cirak, F. 2020.
\newblock Topologically robust CAD model generation for structural
  optimisation.
\newblock \emph{Computer Methods in Applied Mechanics and Engineering}, 369:
  113102.

\bibitem[{Yin et~al.(2021)Yin, Zhang, Wang, Niklaus, Mai, Chen, and
  Shen}]{Yin_2021_CVPR}
Yin, W.; Zhang, J.; Wang, O.; Niklaus, S.; Mai, L.; Chen, S.; and Shen, C.
  2021.
\newblock Learning To Recover 3D Scene Shape From a Single Image.
\newblock In \emph{Proceedings of the IEEE/CVF Conference on Computer Vision
  and Pattern Recognition (CVPR)}, 204--213.

\bibitem[{Zhou et~al.(2023)Zhou, Li, Li, Lange, and Ma}]{featuredcad}
Zhou, T.; Li, H.; Li, X.; Lange, C.~F.; and Ma, Y. 2023.
\newblock Feature-based modeling for variable fractal geometry design
  integrated into CAD system.
\newblock \emph{Advanced Engineering Informatics}, 57: 102006.

\end{thebibliography}

\end{document}